\documentclass[letterpaper]{article} % DO NOT CHANGE THIS
\usepackage{aaai24}  % DO NOT CHANGE THIS
\usepackage{times}  % DO NOT CHANGE THIS
\usepackage{helvet}  % DO NOT CHANGE THIS
\usepackage{courier}  % DO NOT CHANGE THIS
\usepackage[hyphens]{url}  % DO NOT CHANGE THIS
\usepackage{graphicx} % DO NOT CHANGE THIS
\usepackage{natbib}  % DO NOT CHANGE THIS AND DO NOT ADD ANY OPTIONS TO IT
\usepackage{caption} % DO NOT CHANGE THIS AND DO NOT ADD ANY OPTIONS TO IT
\usepackage{algorithm}
\usepackage{algorithmic}
\usepackage{amsmath}
\usepackage{amsfonts}
\usepackage{subcaption} 
\usepackage{multirow}
\usepackage{xcolor} 
\usepackage{color, colortbl}
\usepackage{booktabs}

\usepackage{newfloat}
\usepackage{listings}
\DeclareCaptionStyle{ruled}{labelfont=normalfont,labelsep=colon,strut=off} % DO NOT CHANGE THIS
\floatstyle{ruled}
\newfloat{listing}{tb}{lst}{}
\floatname{listing}{Listing}
\title{DNOI-4DRO: Deep 4D Radar Odometry with Differentiable Neural-Optimization Iterations}
\author{
    Shouyi Lu\equalcontrib, Huanyu Zhou\equalcontrib, Guirong Zhuo\thanks{Corresponding author.}, Xiao Tang
}
\affiliations{
    School of Automotive Studies, Tongji University\\
    \{2210803, 2231634, zhuoguirong, 2433055\}@tongji.edu.cn
}

\usepackage{bibentry}
\begin{document}
	
	\maketitle
	
	\begin{abstract}
		A novel learning-optimization-combined 4D radar odometry model, named DNOI-4DRO, is proposed in this paper. The proposed model seamlessly integrates traditional geometric optimization with end-to-end neural network training, leveraging an innovative differentiable neural-optimization iteration operator. 
		In this framework, point-wise motion flow is first estimated using a neural network, followed by the construction of a cost function based on the relationship between point motion and pose in 3D space. The radar pose is then refined using Gauss-Newton updates. 
		Additionally, we design a dual-stream 4D radar backbone that integrates multi-scale geometric features and clustering-based class-aware features to enhance the representation of sparse 4D radar point clouds. 
		Extensive experiments on the VoD and Snail-Radar datasets demonstrate the superior performance of our model, which outperforms recent classical and learning-based approaches. Notably, our method even achieves results comparable to A-LOAM with mapping optimization using LiDAR point clouds as input.
		Our models and code will be publicly released.
	\end{abstract}
	\section{Introduction}
	Odometry is critical to autonomous driving systems, particularly in GPS-denied environments. It involves using consecutive images or point clouds to estimate the relative pose transformation between two frames. Most existing odometry methods focus primarily on 2D visual odometry \cite{droid,deeppatch,multi,mambavo,leap} or 3D LiDAR \cite{pwclo,lonet,efficientlo,translo,delo,nerfloam,dslo}. However, the inherent characteristics of cameras and LiDARs make these methods vulnerable to challenging weather conditions such as rain, snow, or fog, compromising their robustness in these scenarios.
	
	4D millimeter-wave radar, as an emerging automotive sensor, presents distinct advantages, including robustness under challenging weather and illumination conditions, the ability to measure object velocities and cost-effectiveness \cite{ratrack,milliflow,radarocc,rcbevdet,transloc4d,talk2radar}. These advantages have driven extensive research into 4D Radar Odometry (4DRO). Previous 4D radar odometry works can be classified into two categories: classical methods \cite{4dradarslam,li20234d,efear} and learning-based methods \cite{4drvonet,4dronet}. Classical 4DRO systems typically consist of point cloud association and nonlinear optimization. Nonlinear optimization is critical in enhancing accuracy by integrating continuous sensor observations to optimize long-range trajectories. However, these methods have limited robustness because of the extracted features' poor quality and low resolution \cite{4dradarslam,li20234d}.
	With the development of deep learning, some methods \cite{4dronet,cao} attempt to utilize learning-based methods for 4D radar pose estimation. These methods extract features from two point clouds using a feature extraction network, perform point cloud matching through a feature association network, and finally estimate the pose using a decoding network. Although this direct pose regression method, which does not rely on explicit optimization, can sometimes be more robust, its accuracy often shows limitations across various environments.
	
	To address these problems, we propose DNOI-4DRO, a novel learning-optimization combined 4D radar odometry model. DNOI-4DRO integrates traditional geometric optimization into end-to-end neural network training through an innovatively designed differentiable neural-optimization iteration operator, leveraging the strengths of both classical methods and deep networks. Specifically, we estimate the point-wise motion flow using a neural network, construct the cost function based on the motion and pose relationships of points in 3D space, and update the radar pose using the Gauss-Newton to maximize its compatibility with the current point motion flow estimate. Furthermore, to extract robust 4D radar point features, we propose a novel feature extraction network that enhances the representation of radar point clouds by integrating multi-scale geometric features with clustering-based class-aware features, thereby improving motion estimation in complex environments. We perform extensive evaluation across two datasets, demonstrating state-of-the-art performance in all cases. To summarize, our main contributions are as follows:
	
	(1) We propose DNOI-4DRO, a novel framework for 4D radar odometry. By exploiting the relationship between point motion and pose, DNOI-4DRO couples a geometric optimizer with a neural network to construct a fully end-to-end architecture. (2) We specially design an efficient radar feature extractor comprising a dual-stream 4D radar backbone, which extracts radar features with two representations. 
	(3) Finally, our method is demonstrated on VoD \cite{vod} and Snail-Radar \cite{snailradar} datasets. Our method outperforms all classical and learning-based 4D radar odometry methods, achieving competitive results with A-LOAM, which uses LiDAR point clouds as input.
	\section{Related Work}
	\subsection{Classical 4D Radar Odometry}
	Due to the similarity in spatial representation between 4D radar and LiDAR point clouds, most existing classical 4D radar odometry methods either directly adopt LiDAR odometry methods or modify them accordingly. Iterative Closest Point (ICP) is the most widely used LiDAR odometry method, computing relative pose transformations by minimizing the Euclidean distance between two point clouds. Based on error measurement, ICP can be categorized into ICP-point2point \cite{p2picp} and ICP-point2plane \cite{p2plicp}, aiming to shorten point-to-point and point-to-plane distances. Generalized ICP (GICP) \cite{gicp} further intends to combine the advantages of both. Building on GICP, 4DRadarSLAM \cite{4dradarslam} models the spatial probabilistic distribution of radar points based on uncertainties in range and azimuth angle measurements. It proposes an Adaptive Probability Distribution-GICP (APDGICP) method based on the modeling results. RIV-SLAM \cite{rivslam} further refines this by considering the impact of the angle of arrival on radar measurement uncertainty to achieve a more accurate spatial probabilistic distribution of radar points. In addition to ICP, the Normal Distribution Transform (NDT) \cite{ndt} is a widely used LiDAR odometry method. NDT represents point clouds with Gaussian distributions, converting pose estimation into an optimization problem that minimizes the Gaussian distribution error between the source and target point clouds. Expanding upon NDT, Li et al. \cite{li20234d} integrate the measurement uncertainties of 4D radar points into the calculations for the mean and variance of the normal distribution, aiming to reduce degradation effects caused by sparse 4D radar point clouds. Additionally, 4D iRIOM \cite{4driom} introduces a point cloud registration method based on a distribution-to-multi-distribution distance metric, effectively addressing the sparsity issue of 4D radar point clouds and enhancing pose estimation robustness.
	
	Our method draws on the cost function modeling approach of the ICP-point2point \cite{p2picp}, a simpler optimization problem that avoids extracting complex representations like line and plane features. Instead of relying on nearest-neighbor queries to find matching points, our method uses a neural network to estimate the point motion field. This approach mitigates the challenging non-bijective correspondence between two point clouds caused by the sparsity and noise inherent in 4D radar point clouds.
	
	\subsection{Deep 4D Radar Odometry}
	Deep learning has made significant strides in visual and LiDAR odometry, yet 4D radar odometry remains a challenging problem. 4DRONet \cite{4dronet} proposes a sliding-window-based hierarchical optimization method to estimate and refine poses in a coarse-to-fine manner. SelfRO \cite{10422466} introduces a self-supervised 4D radar odometry method that uses radar point velocity information to construct a novel velocity-aware loss function, effectively guiding network training. CMFlow \cite{cmflow} introduces a cross-modal supervision approach for estimating 4D radar scene flow, simultaneously providing pose estimation as an intermediate output. CAO-RONet \cite{cao} introduces a sliding-window-based optimizer that leverages historical priors to enable coupled state estimation and correct inter-frame matching errors. Although these methods show promising results, they often suffer from limited accuracy across diverse environments, and their interpretability remains constrained.
	
	Moreover, we are inspired by the ``Differentiable Recurrent Optimization-Inspired Design'' proposed by DROID-SLAM \cite{droid} for visual odometry. The work combines iterative visual correspondence updates with differentiable bundle adjustments to optimize pose estimation. This concept has also been applied to optimizing pixel-level 3D motion \cite{3draft}. However, both works are based on RAFT \cite{raft}, which performs optical flow updates and optimizes variables in 2D space. This paper introduces the first neural-optimization module for 4D radar-based 3D point cloud odometry by tightly coupling RAFT/DROID-style differentiable iterations with Bundle Adjustment.

	\section{DNOI-4DRO}
	\label{sec:method}
\begin{figure*}[t]
		\centering\includegraphics[width=0.99\linewidth]{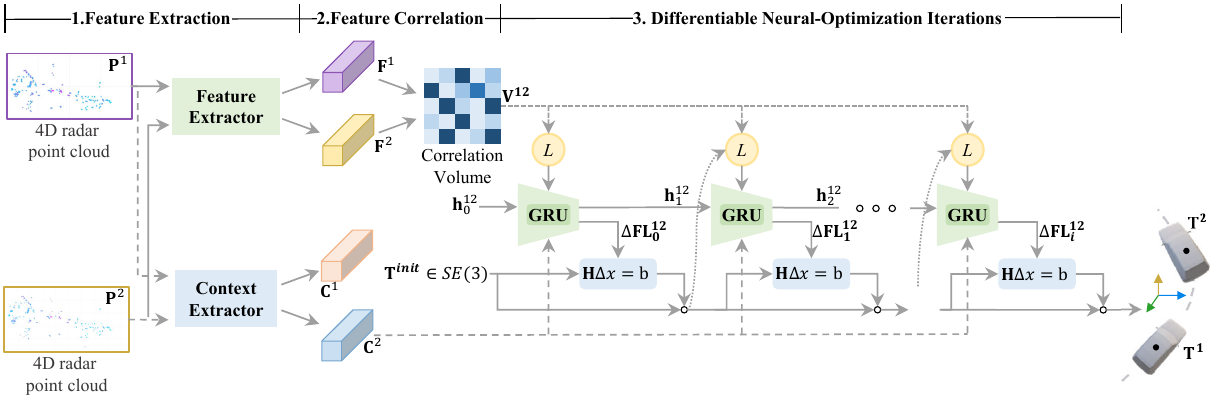} 
		\caption{Overview of our backbone. (1) The feature and context extractors encode the point and context features of the input point cloud, respectively. (2) The feature correlation module constructs an all-pair correlation volume by calculating the matrix dot product of two-point features. (3) In each iteration, the differentiable neural-optimization iteration operator uses the pose estimated in the previous iteration to look up correlation features from the correlation volume, which are then processed through a GRU \cite{gru} to generate a point motion field. The point motion field is fed into a least-squares-based optimization layer, where the pose is updated based on geometric constraints. After multiple iterations, the network outputs the predicted pose.}
		\label{fig1}
	\end{figure*}

	We propose a backbone network for pose estimation between two 4D radar point clouds and then construct a 4DRO system upon it. We provide an overview in Figure~\ref{fig1}. The backbone network firstly extracts robust point and context features for each 4D radar point cloud. Then, the feature correlation volume is constructed using point features. Next, the differentiable neural-optimization iteration operator is introduced for pose estimation and refinement. Finally, the network outputs the pose $\mathbf{T}^1\in\mathbb{R}^{4\times4}$ and $\mathbf{T}^2\in\mathbb{R}^{4\times4}$ of the two 4D radar point clouds in the world coordinate system.
	
	\subsection{Feature Extraction}
	\label{sec:fe}
\begin{figure}[t]
		\centering
		\includegraphics[width=0.45\textwidth]{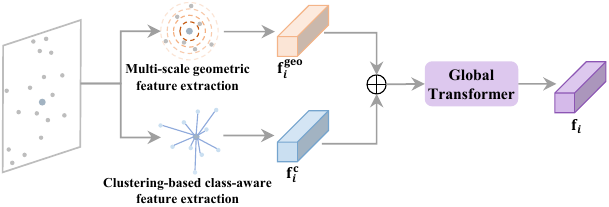}\vspace{-0.2cm}
		\caption{The structure of dual-stream radar feature extraction network.}
		\label{fig2}
	\end{figure}

	We employ two radar feature extraction networks with identical architectures but distinct functions to extract the point and context features from the 4D radar point cloud. The context features are provided as input to the recurrent iteration operator, whereas the point features are used to evaluate the point similarity using a dot product. To obtain fine-grained 4D radar point features, we propose a dual-stream 4D radar encoding network that integrates multi-scale geometric feature extraction, clustering-based class-aware feature extraction, and a global Transformer module, as illustrated in Figure~\ref{fig2}. This network offers a more comprehensive feature representation for each 4D radar point, thereby significantly improving the robustness of odometry across diverse environments.
	
	\subsubsection{Multi-scale Geometric Feature Extraction}
	Given the 4D radar point clouds $\mathbf{P} = \{ \mathbf p_i|\mathbf p_i \in {\mathbb{R}^3}\} _{i = 1}^N$, we employ four parallel Set Abstraction (SA) layers \cite{pointnet++} to encode the multi-scale geometric features. Each SA layer uses a different grouping radius to address the issue of non-uniform density in radar point clouds. At each scale $s$, each radar point $\mathbf p_i$ generates a local geometric feature $\mathbf{f}_{i,s}^{\text{geo}}$. The geometric features across all scales are concatenated to form the multi-scale geometric feature representation of the 4D radar point, $\mathbf{f}_{i}^{\text{geo}}=\text{concat}(\{\mathbf{f}_{i,s}^{\text{geo}}\}_{s=0}^3)$.
	\subsubsection{Clustering-based Class-aware Feature Extraction}
	To address the inherently non-uniform spatial distribution of 4D radar point clouds, we propose a novel clustering-based class-aware 4D radar point feature extraction method, consisting primarily of three components: 4D radar point clustering, class-aware feature aggregation, and feature dispatching.
	
	\noindent{}{\bf 4D Radar Point Clustering:} We first use PointNet \cite{pointnet} to map the 4D radar points to a high-dimensional feature space $\mathbf{f}_{i}^{\text{p}}$ for similarity computation. Next, we apply Farthest Point Sampling (FPS) \cite{pointnet++} to evenly select $C$ center points in space, and the center feature $\mathbf{f}^{\text{ct}}$ is computed by averaging its $K$ nearest points. We then calculate the pairwise cosine similarity matrix $\mathbf S\in\mathbb{R}^{C\times N}$ between each point and the set of center points. After that, we allocate each point to the most similar center, resulting in $C$ clusters.
	
	\noindent{}{\bf Class-aware Feature Aggregation:} Following \cite{ma2023image}, we dynamically aggregate all radar point features within the same cluster based on the similarities to the cluster center. Given the cluster which contains $M$ radar points around the $j$-th cluster center, the aggregated feature $\mathbf{f}_{j}^{\text{a}}$ is calculated by:
	\begin{align}
		\mathbf f_j^\text{a} &= \frac{1}{\mathbf A} \left(\mathbf f_j^{\text{ct}} + \sum_{m=1}^{M} \text{sigmoid}(\alpha \mathbf s_{jm} + \beta) \cdot \mathbf f_m^\text p \right), \\
		\mathbf A &= 1 + \sum_{m=1}^{M} \text{sigmoid}(\alpha \mathbf s_{jm} + \beta),
	\end{align}
	where $\mathbf f_j^{\text{ct}}$ represents the feature of the $j$-th cluster center, and $\mathbf s_{jm}$ denotes the similarity score between the $m$-th radar point and the $j$-th cluster center.
	$\alpha$ and $\beta$ are learnable scalars used to scale and shift the similarity, while $\text{sigmoid}(\cdot)$ is the sigmoid function that rescales the similarity to $\text{(0,1)}$. $\mathbf A$ is the normalization factor.
	
	\noindent{}{\bf Feature Dispatching:} After obtaining the class-aware aggregated features, we employ an adaptive process to assign each point within a cluster based on its similarity, resulting in a more coherent and representative understanding of the overall structure and context within the cluster. For each point embedding $\mathbf f_m^{\text{p}}$, the updated point embedding $\mathbf f_m^{\text{c}}$ is computed using the following formula:
	\begin{equation}\mathbf f_m^{\text{c}} = \mathbf f_m^{\text{p}} + \text{sigmoid}(\alpha \mathbf s_{jm} + \beta) \cdot \mathbf f_j^\text{a},\end{equation}
	We concatenate the multi-scale geometric features of radar points with the class-aware features to construct the geometric-class joint feature representation $\mathbf f_i^\text{l}$.
	\subsubsection{Global Transformer}
	The global transformer employs an attention mechanism \cite{attention} across the entire radar point cloud to capture the long-range correlation of points. For each point $\mathbf p_i \in \mathbf{P}$, the global transformer applies an attention mechanism to all other points $\mathbf p_j \in \mathbf{P}$:
	\begin{equation}\mathbf f_i = \sum_{\mathbf f_j^\text{l} \in \mathcal{X}^\text{l}} \langle \alpha^g(\mathbf f_i^\text{l}), \beta^g(\mathbf f_j^\text{l}) \rangle \gamma^\text{l}(\mathbf f_j^\text{l}),
	\end{equation}
	where $\mathcal{X}^\text{l}$ represents the geometric-class joint feature set, $\alpha^g(\cdot)$, $\beta^g(\cdot)$, and $\gamma^g(\cdot)$ are shared learnable linear transformations. $\langle\cdot\rangle$ denotes a scalar product. Linear layer, layer norm, and skip connection are further applied to complete the global transformer module.
	\subsection{Feature Correlation}
	\label{sec:corr}
	We construct a feature correlation volume by evaluating feature similarity between all point pairs. Given point features $\mathbf F^1 = \{ \mathbf f_i^1|\mathbf f_i^1 \in {\mathbb{R}^D}\} _{i = 1}^N$ and $\mathbf F^2 = \{ \mathbf f_j^2|\mathbf f_j^2 \in {\mathbb{R}^D}\} _{j = 1}^N$, where $D$ is the feature dimension, the correlation volume $\mathbf V^{12}\in\mathbb{R}^{N\times N}$ can be computed through matrix dot product:
	\begin{equation}\mathbf V^{12} = \mathbf F^1 \cdot (\mathbf F^2)^T.\end{equation}
	\subsection{Differentiable Neural-Optimization Iteration Operator}
	\label{sec:uo}
\begin{figure}[t]
		\centering
		\includegraphics[width=0.45\textwidth]{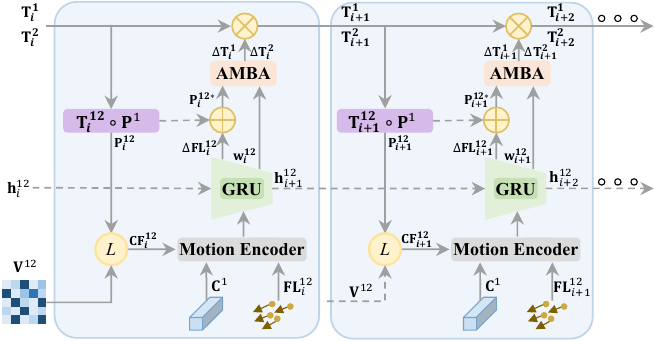}\vspace{-0.2cm}
		\caption{Illustration of the differentiable neural-optimization iteration operator, which predicts point motion flow revisions and maps them to pose updates through the AMBA layer.}
		\label{fig3}
	\end{figure}

	The core of our approach is the differentiable neural-optimization iteration operator, which contains three key parts, as illustrated in Figure.~\ref{fig3}: (1) an adaptive patch-to-patch correlation lookup method that retrieves correlation features from the precomputed correlation volume; (2) a GRU \cite{gru} that predicts updates to the point motion flow along with associated confidence weights; and (3) an All-point Motion-only Bundle Adjustment layer (AMBA) computes pose that aligns with the newly predicted point motion flow and confidence.
	\subsubsection{Adaptive Patch-to-Patch Correlation Look-up Layer}
	At the start of each iteration, we use the pose estimates from the previous iteration to calculate the correspondences. Specifically, for the $i$-th iteration, given the point clouds $\mathbf{P}^1$ and $\mathbf{P}^2$, along with their respective poses $\mathbf{T}^1_{i}$ and $\mathbf{T}^2_{i}$, we compute the all-point correspondence field $\mathbf{P}^{12}_{i}\in\mathbb{R}^{N\times3}$:
	\begin{equation}\mathbf{P}^{12}_{i} = \mathbf{T}^{12}_{i} \cdot \mathbf{P}^1, \quad \mathbf{T}^{12}_{i} = \mathbf{T}^2_{i} \cdot (\mathbf{T}^1_{i})^{-1}, \end{equation}
	$\mathbf{P}^{12}_{i}$ represents the coordinates of points $\mathbf{P}^1$ mapped into $\mathbf{P}^2$ using the estimated pose. Then, we incorporate the Patch-to-Patch correlation concept from HALFlow \cite{wang2021hierarchical} into the point correlation look-up mechanism of PV-RAFT \cite{pvraft} to learn fine-grained correlation features of sparse 4D radar points, thereby guiding the right flowing direction of point motion. Specifically, for each point $\mathbf p_{i,j}^{12}$ in $\mathbf{P}^{12}_{i}$, we first retrieve the initial correlation features from the correlation volume using nearest neighbor queries. Subsequently, based on the 3D local geometric structure of $\mathbf p_{i,j}^{12}$ and its neighboring points in $\mathbf{P}^{12}_{i}$, we adaptively weight the initial correlation features of these neighbors to obtain the final correlation feature $\mathbf{cf}_{i,j}^{12}$.
	\subsubsection{Point Motion Flow Update and Confidence Weight Prediction}
	We encode motion features using the correlation features $\mathbf{CF}^{12}_i= \{ \mathbf{cf}_{i,j}^{12}|\mathbf{cf}_{i,j}^{12} \in {\mathbb{R}^D}\} _{j = 1}^N$, the context features $\mathbf C^1$, and the flow estimate $\mathbf{FL}^{12}_i \in \mathbb{R}^{N\times3}$ from the previous iteration:
	\begin{equation}
		\mathbf{MF}^{12}_i = \text{concat}(\mathbf{CF}^{12}_i, \mathbf{C}^1, \text{MLP}(\mathbf{FL}^{12}_i)).
	\end{equation}
	The motion features $\mathbf{MF}^{12}_i$ is used as the input to the GRU \cite{gru} to update the current hidden state $\mathbf{h}^{12}_i$. Instead of predicting updates to the pose directly, we predict updates in the space of point motion flow fields. We map the updated hidden state through two additional MLP layers to produce two outputs: (1) a revision flow field $\Delta \mathbf{FL}^{12}_i \in \mathbb{R}^{N\times3}$ and (2) associated confidence $\mathbf{w}^{12}_i \in \mathbb{R}_+^{N\times3}$. The revision $\Delta \mathbf{FL}^{12}_i$ is a correction term predicted by the network to correct errors in the all-point correspondence field. We denote the corrected correspondence as $\mathbf{P}^{12*}_i=\mathbf{P}^{12}_i+\Delta \mathbf{FL}^{12}_i$. The confidence term is further processed with a sigmoid operation to constrain its values within the $(0,1)$. This term mitigates the adverse impact of inaccurate point motion flow estimation caused by occlusion, moving objects and noisy points on pose estimation accuracy.
	\subsubsection{All-point Motion-only Bundle Adjustment Layer}
	The All-point Motion-only Bundle Adjustment Layer (AMBA) maps flow revisions into pose updates. The primary objective of the AMBA layer is to minimize the discrepancy between the transformed point cloud and the corrected correspondence $\mathbf{P}^{12*}_i$. Specifically, we define the cost function as:
	\begin{equation}
		E(\mathbf{T}^1, \mathbf{T}^2) = \left\| \mathbf{P}^{12*}_i - \mathbf{T}^2 \cdot (\mathbf{T}^1)^{-1} \cdot \mathbf{P}^{1} \right\|^2_{\Sigma_{12}},
	\end{equation} 
	where $\left\|\cdot\right\|^2_{\Sigma}$ is the Mahalanobis distance which weights the error terms based on the confidence weights $\mathbf{w}^{12}_i$. In this paper, we optimize the pose using the Gauss-Newton algorithm, based on the special Euclidean Lie algebra $\text{se}(3) = \left\{\xi = \left( \begin{array}{c} \rho \\ \phi \end{array} \right), \ \rho \in \mathbb{R}^3, \ \rho \in \text{so}(3) \right\}$. First, we compute the Jacobian matrices $\mathbf{J}^1 \in \mathbb{R}^{N\times3\times6}$ and $\mathbf{J}^2 \in \mathbb{R}^{N\times3\times6}$ of the cost function with respect to $\mathbf{T}^1$ and $\mathbf{T}^2$ using the perturbation model. 
	\begin{align}
		\mathbf{J}^1 &= \left[ \begin{array}{cc} \mathbf I & -(\mathbf R^{12} \circ \mathbf P^1 + \mathbf t^{12})^{\wedge} \\ \mathbf 0^T & \mathbf 0^T \end{array} \right] \circ \text{Adj}_{\mathbf T^{12}}, \\
		\mathbf{J}^2 &= - \left[ \begin{array}{cc} \mathbf I & -(\mathbf R^{12} \circ \mathbf P^1 + \mathbf t^{12})^{\wedge} \\ \mathbf 0^T & \mathbf 0^T \end{array} \right],
	\end{align}
	where $\left( \cdot \right)^{\wedge}$ denotes the antisymmetric matrix operation and $\text{Adj}_{\left( \cdot \right)}$ denotes the adjoint operator.
	
	Then, we linearizes the cost function, and solves for the update:
	\begin{equation}
		\mathbf{J}^T \text{diag}(\mathbf{w}^{12}_i) \mathbf{J} \Delta \xi = -\mathbf{J}^T E(\mathbf T^1, \mathbf T^2),
		\label{eq10}
	\end{equation}
	where $\Delta \xi=(\Delta \xi^1,\Delta \xi^2)$ represents the pose update in the Lie algebra. Eqn.\ref{eq10} can be rewritten as a linear system:
	\begin{equation}
		\mathbf H \Delta \xi = \mathbf b, \quad \mathbf H = \mathbf J^T \text{diag}(\mathbf{w}^{12}_i) \mathbf J, \quad \mathbf b = -\mathbf J^T E(\mathbf T^1, \mathbf T^2)
		.
		\label{eq11}
	\end{equation}
	This linear system is solved using Cholesky decomposition to obtain the updates $\Delta \xi^1$ and $\Delta \xi^2$. The updates refine the pose estimates from the previous iteration, and the updated poses serve as input for the next neural-optimization iteration module:
	\begin{equation}
		\mathbf{T}^1_{i+1} = \text{Exp} \left( \Delta \mathbf{\xi}^1 \right) \cdot \mathbf{T}^1_i,\quad 
		\mathbf{T}^2_{i+1} = \text{Exp} \left( \Delta \mathbf{\xi}^2 \right) \cdot \mathbf{T}^2_i.
	\end{equation}
	Based on the properties of the Jacobian matrix and Cholesky decomposition, gradients are backpropagated using the chain rule. Consequently, the AMBA layer functions as a differentiable optimization layer, enabling end-to-end training of the entire architecture during the training process.
	\section{Training and Implementation}\label{sec:tas}
	DNOI-4DRO is implemented in PyTorch, with the LieTorch extension \cite{teed2021tangent} used to perform backpropagation in the tangent space of all group elements.
	\subsection{Constructing Frame Graph}
	Each training example consists of a 7-frame 4D radar point cloud sequence. For each point cloud frame in the training samples, edges are constructed with its three nearest neighboring frames to form a frame graph. In this graph, any two frames connected by an edge undergo pose estimation using the backbone network. Additionally, to remove the 6-dof gauge freedom, the first pose in each training sample is fixed to the ground truth.
	\subsection{Supervision}
	We supervise our network using a pose loss. Given a set of ground truth poses $\{\tilde{\mathbf{T}}\}_{i=1}^L$ and predicted poses $\{\mathbf{T}\}^L_{i=1}$, the pose loss is taken to be the distance between the ground truth and predicted poses:
	\begin{equation}
		L_{pose} = \sum_{i=1}^L \left\| \text{Log}_{SE3}\left((\tilde{\mathbf{T}}^i)^{-1} \cdot \mathbf{T}^i \right) \right\|^2,
	\end{equation}
	where $\text{Log}_{SE3}(\cdot)$ represents the logarithmic map from Lie group to Lie algebra.
	\subsection{Training Details}
	We train for 30 epochs on an A100 GPU with a batch size of 4. We use the Adam optimizer and start with an initial learning rate of $2e^{-4}$, which decays by a factor of 0.1 every 10 epochs. $N$ is set to be 512 in the proposed network. The radar point heights are constrained within the range of $[-2m, 10m]$ to retain reliable points. During training, random transformation matrices are applied to the points and the ground truth pose to increase data diversity. We select 8 cluster centers. The iteration operator is unrolled for 15 iterations during training, with two bundle adjustment steps performed within each iteration. The ground truth pose of the first frame in each training sequence serves as the initial pose for all radar frames in that sequence, with the initial point motion flow set to zero.
	\subsection{4DRO System}
	We apply the backbone network during inference to construct a fully 4DRO system that processes a 4D radar point cloud stream for real-time localization.
	
	\noindent{}{\bf Initialization:}
	We initialize with 8 frames, continuously adding new frames until 8 frames are accumulated, after which we perform 12 iterations of the iteration operator.
	
	\noindent{}{\bf Tracking:}
	We maintain a frame graph. After initialization, when a new frame is added, we extract features and add the new frame to the frame graph, establishing edges with all frames within an index distance of 2. The pose is initialized using a linear motion model. We then run four iterations of the iteration operator, performing two bundle adjustment iterations within each iteration. We fix the initial frame's pose in the frame graph to remove gauge freedom. After the new frame is tracked, we remove the oldest frame.
	\section{Experiments}
	\label{sec:exp}
	We evaluate the method on two 4D radar odometry benchmarks, with dataset details in the appendix. We follow protocols of DVLO \cite{dvlo} to evaluate our method with two metrics: (1) Average sequence translational RMSE (\%). (2) Average sequence rotational RMSE ($^{\circ}$/100m).
	\subsection{Performance Evaluation}
	\noindent{}{\bf VoD Results:} 
	\begin{table*}[t]
		\centering
		\footnotesize
		\begin{center}
			\resizebox{1.0\textwidth}{!}
			{	
				\begin{tabular}{cc||cc|cc|cc|cc|cc|cc|cc||cc}
					\toprule
					\multicolumn{2}{c||}{\multirow{2}{*}{Method}} & \multicolumn{2}{c|}{03} & \multicolumn{2}{c|}{04} & \multicolumn{2}{c|}{09} & \multicolumn{2}{c|}{17} & \multicolumn{2}{c|}{19} & \multicolumn{2}{c|}{22} & \multicolumn{2}{c||}{24} & \multicolumn{2}{c}{Mean} \\ \cmidrule(l){3-18} 
					%\hline{2-25}
					%\cline{2-17}\noalign{\smallskip}
					
					%\multirow{-2}{*}{\begin{tabular}[c]{@{}c@{}}Method \end{tabular}}
					& &  $t_{rel}$  & $r_{rel}$   & $t_{rel}$   & $r_{rel}$               & $t_{rel}$                          & $r_{rel}$   & $t_{rel}$ & $r_{rel}$   & $t_{rel}$                          & $r_{rel}$   & $t_{rel}$ & $r_{rel}$    & $t_{rel}$                          & $r_{rel}$   & $t_{rel}$ & $r_{rel}$ \\
					%\midrule
					%\cline{1-25}\noalign{\smallskip}
					\hline\hline
					\noalign{\smallskip}
					
					%\noalign{\smallskip}
					\multirow{4}{*}{Classical-based methods}
					&ICP-po2po    
					& 0.39  & 1.00    
					& 0.21  & 1.14    
					& 0.15  & 0.72    
					& 0.16  & 0.53    
					& 1.40  & 4.70    
					& 0.44  & 0.76   
					& 0.24  & 0.77  
					& 0.427  & 1.374
					\\ 
					
					& ICP-po2pl     
					& 0.42 & 2.19    
					& 0.37 & 1.83    
					& 0.50 & 1.32    
					& 0.23 & 0.68     
					& 3.04 & 5.62    
					& 0.42 & 1.20 
					& 0.35 & 0.67 
					& 0.761 & 1.930
					\\ 
					
					& GICP    
					& 0.46 & 0.68     
					& 0.30 & 0.39     
					& 0.51 & 0.32    
					& 0.40 & 0.10    
					& 0.51 & 1.23    
					& 0.34 & 0.57 
					& 0.15 & 0.30 
					& 0.381 & 0.513 
					\\ 
					& NDT    
					& 0.55 & 1.60    
					& 0.47 & 0.91     
					& 0.46 & 0.56    
					& 0.44 & 0.40    
					& 1.33 & 2.58    
					& 0.47 & 1.10 
					& 0.36 & 1.84 
					& 0.583 & 1.284  
					\\ 
					\cmidrule(r){1-18}
					\multirow{5}{*}{LiDAR-based methods}
					& Full A-LOAM    
					& NA     & NA         
					& 0.03  & 0.09        
					& 0.04  & 0.19    
					& 0.02  & 0.04  
					& 0.38  & 1.35  
					& 0.06  & 0.18    
					& 0.06  & 0.20 
					& 0.098  & 0.342
					\\ 
					
					& A-LOAM w/o mapping    
					& NA     & NA         
					& 0.14  & 0.35         
					& 0.16  & 1.23    
					& 0.09  & 0.26    
					& 1.17  & 4.63    
					& 0.27  & 0.92 
					& 0.16  & 0.81 
					& 0.332  & 1.370
					\\ 
					&PWCLO-Net   
					&0.26  & 0.37     
					&0.31  & 0.40     
					&0.38  & 0.55    
					&0.27  & 0.39  
					&1.23  & 0.91    
					&0.23  & 0.35 
					&0.46  & 0.82
					&0.449  & 0.541
					\\  
					\cmidrule(r){1-18}
					\multirow{4}{*}{4D Radar-based methods}
					&RaFlow    
					& 0.87  & 2.09         
					& 0.07  & 0.44         
					& 0.11  & 0.09  
					& 0.13  & 0.03    
					& 1.22  & 4.09    
					& 0.72  & 1.34 
					&0.25 &1.14 
					&0.481 &1.317 
					\\
					&CMFlow   
					& 0.06 & 0.10     
					& 0.05 & 0.09    
					& 0.09 & 0.14    
					& 0.06 & 0.03    
					& 0.28 & 0.94     
					& 0.14 & 0.29 
					&0.12  &0.58 
					&0.114  &0.310 
					\\    
					&4DRONet   
					&0.08  &0.10	  
					&0.04 &0.07	 
					&0.13 &0.38
					&0.09 &0.10  
					&0.91 &0.62
					&0.23 &0.32
					&0.28 &1.20
					&0.251 &0.398
					\\ 
					&CAO-RONet   
					&0.049  &0.034	  
					&0.042 &0.040	 
					&0.064 &0.089
					&0.102 &\bf0.013
					&0.024 &0.045
					&0.082 &0.063
					&0.135 &0.076
					&0.071 &0.051
					\\
					\cmidrule(r){1-18}       					
					\multicolumn{2}{c||}{Ours}      
					&\bf0.021 & \bf0.018
					&\bf0.022 & \bf0.026
					&\bf0.017 & \bf0.016
					&\bf0.018 & 0.021
					&\bf0.021 & \bf0.038
					&\bf0.025 & \bf0.024
					&\bf0.034 & \bf0.048
					&\bf0.023({$\downarrow$67.6\%}) & \bf0.027({$\downarrow$47.1\%})   
					\\ \bottomrule
				\end{tabular}
			}
		\end{center}
		\caption{The 4D radar odometry experiment results on the VoD dataset \cite{vod}. $t_{rel}$ and $r_{rel}$ denote the average translational RMSE (m/m) and rotational RMSE ($^{\circ}$/m), respectively, on all possible subsequences in the length of $20,40,...,160m$. All methods listed use 4D radar point clouds as input. Full A-LOAM is a complete SLAM system and others only include odometry. The best results are bold.}	
		\label{table:vod1}
	\end{table*}
	\begin{table*}[t]
		\centering
		\footnotesize
		\begin{center}
			\resizebox{1.0\textwidth}{!}
			{
				\begin{tabular}{l|c||cc|cc|cc|cc|cc|cc|cc||cc}
					\toprule
					& &  \multicolumn{2}{c|}{03}  &\multicolumn{2}{c|}{04}      & \multicolumn{2}{c|}{09} & \multicolumn{2}{c|}{17} &  \multicolumn{2}{c|}{19} & \multicolumn{2}{c|}{22} & \multicolumn{2}{c||}{24}  &\multicolumn{2}{c}{Mean} \\ 
					%\hline{2-25}
					\cline{3-18}\noalign{\smallskip}
					
					\multirow{-2}{*}{Method} & \multirow{-2}{*}{Input}
					&  $t_{rel}$  & $r_{rel}$   & $t_{rel}$                       & $r_{rel}$               & $t_{rel}$                          & $r_{rel}$   & $t_{rel}$ & $r_{rel}$   & $t_{rel}$                          & $r_{rel}$   & $t_{rel}$ & $r_{rel}$    & $t_{rel}$                          & $r_{rel}$   & $t_{rel}$ & $r_{rel}$ \\
					
					\hline\hline
					\noalign{\smallskip}
					
					Full A-LOAM 
					&L    
					&0.05	&0.11
					&0.03	&0.10
					&0.03	&0.08
					&0.02	&\bf0.02
					&0.05	&0.14
					&0.03	&0.09
					&\bf0.03	&0.05
					&0.034	&0.084
					
					\\ 
					
					A-LOAM w/o mapping
					&L    
					&0.06	&0.10
					&0.03	&0.12
					&0.06	&0.06
					&0.03	&\bf0.02
					&0.14	&0.19
					&0.06	&0.11
					&0.06	&0.08
					&0.063	&0.097
					
					\\                		
					
					4DRVONet
					& C+R      
					&\bf0.02 & 0.02
					&\bf0.01 & \bf0.02
					&0.03 & 0.05
					&0.11 & 0.03
					&0.26 & 0.08
					&0.06 & 0.09
					&0.09 & 0.18
					&0.083 & 0.067
					\\
					%\gr
					Ours
					& R      
					&0.021 & \bf0.018
					&0.022 & 0.026
					&\bf0.017 & \bf0.016
					&\bf0.018 & 0.021
					&\bf0.021 & \bf0.038
					&\bf0.025 & \bf0.024
					&0.034 & \bf0.048
					&\bf0.023({$\downarrow$32.4\%}) & \bf0.027({$\downarrow$59.7\%})  
					\\ \bottomrule
				\end{tabular}
			}
		\end{center}
		\caption{The experiment results on the VoD dataset \cite{vod} using methods with LiDAR or 4D radar-camera inputs. `L', `C', and `R' represent LiDAR, camera, and 4D radar, respectively. The best results are bold.} 
		\vspace{-3mm}
		\label{table:vod2}
	\end{table*}
	Quantitative results are listed in Table~\ref{table:vod1}. ICP-point2point (ICP-po2po) \cite{p2picp}, ICP-point2plane (ICP-po2pl) \cite{p2plicp}, GICP \cite{gicp}, NDT \cite{ndt} are several classical point cloud odometry methods. TransLO \cite{translo} and EfficientLO \cite{efficientlo} rely on the imaging principle of LiDAR to project point clouds onto a 2D cylindrical surface as network input. However, 4D radar operates with a different imaging mechanism and produces only about 512 sparse points per frame, making projection-based approaches unsuitable for such data. Therefore, we primarily compare our method with leading open-source LiDAR odometry techniques that directly process raw 3D point clouds, such as PWCLO-Net \cite{pwclo}.
	A-LOAM \cite{loam} is a conventional LiDAR odometry method that achieves state-of-the-art performance on the KITTI Odometry benchmark \cite{kitti}. We use 4D radar point clouds as input for these methods. The experimental results indicate that, while these methods perform effectively on LiDAR point clouds, their performance deteriorates significantly on 4D radar point clouds due to the sparse, noisy, and non-panoramic characteristics of 4D radar data. RaFlow \cite{raflow} and CMFlow \cite{cmflow} are 4D radar-based scene flow estimation networks incorporating odometry estimation as an intermediate task. 4DRONet \cite{4dronet} and CAO-RONet \cite{cao} are odometry networks specifically tailored for processing 4D radar data. Our method outperforms all previous 4D radar odometry methods due to its innovative architecture and robust point cloud feature extraction. Notably, our method reduces mean translation and rotation errors by 67.6\% and 47.1\%, respectively, compared to the previous best learning-based method.
	
	Table~\ref{table:vod2} shows the evaluation results of Full A-LOAM and A-LOAM without mapping using 64-beam LiDAR point clouds as input, along with the results of the 4D radar-camera fusion odometry method (4DRVONet) \cite{4drvonet}. It can be seen that our method achieves competitive performance with Full A-LOAM in short-range localization. Compared with 4DRVONet, our method achieves a reduction of 72.3\% in mean translation error and 59.7\% in mean rotation error on the test sequences. The experiment results prove the effectiveness of our proposed neural-optimization iteration module and demonstrate its great potential in 4D radar odometry tasks. 
	
	\noindent{}{\bf Snail-Radar Results:} 
	\begin{table}[t]
		\centering
		\footnotesize
		\begin{center}	
			\resizebox{1\columnwidth}{!}
			{
				\begin{tabular}{l|c||cc|cc|cc||cc}
					\toprule
					& & \multicolumn{2}{c|}{if} &\multicolumn{2}{c|}{iaf} &\multicolumn{2}{c||}{st} &\multicolumn{2}{c}{Mean} \\ 
					%\hline{2-25}
					\cline{3-10}\noalign{\smallskip}
					
					\multirow{-2}{*}{Method} & \multirow{-2}{*}{Input}
					& $t_{rel}$  & $r_{rel}$  &  $t_{rel}$  & $r_{rel}$  & $t_{rel}$                       & $r_{rel}$ & $t_{rel}$   & $r_{rel}$      \\
					%\midrule
					%\cline{1-25}\noalign{\smallskip}
					\hline\hline
					\noalign{\smallskip}
					
					Full A-LOAM 
					&L
					&\bf0.92 & \bf0.41 
					&\bf1.54 & \bf0.54
					&\bf{1.93} & \bf2.54
					&\bf1.465  & \bf1.163
					\\ 
					
					A-LOAM w/o mapping 
					&L
					&14.74 & 4.10 
					&14.75 & 3.88
					&12.26 & 9.62
					&13.916  & 5.867
					\\ 
					\cmidrule(r){1-10}
					ICP-po2po
					&R
					&35.54 & 13.98
					&37.09 & 10.18
					&26.31 & 60.07
					&32.980 & 28.077
					\\ 
					ICP-po2pl
					&R
					&27.16 & 11.17
					&30.81 & 11.35
					&20.70 & 39.96
					&26.226 & 20.827
					\\ 
					GICP
					&R
					&31.20 & 11.37
					&31.79 & 9.7
					&26.36 & 51.22
					&29.785 & 24.097
					\\
					\cmidrule(r){1-10}
					4DRadarSLAM
					&R
					&13.32 & 6.11
					&16.13 & 6.62
					&10.70 & 15.43
					&13.386 & 9.387
					\\
					\cmidrule(r){1-10}
					%\gr
					Ours   
					&R   
					& \underline{3.92} & \underline{1.85}
					& \underline{6.10} & \underline{2.21}
					& \underline{3.76} & \underline{3.94}
					& \underline{4.598} & \underline{2.670}
					\\ \bottomrule
				\end{tabular}
			}
		\end{center}
		\caption{The experiment results on the Snail-Radar dataset \cite{snailradar}. $t_{rel}$ and $r_{rel}$ denote the average translational RMSE (\%) and rotational RMSE ($^{\circ}$/100m), respectively, on all possible subsequences in the length of $100,200,...,800m$. `L' and `R' represent LiDAR and 4D radar, respectively. The best results are shown in bold, and the second-best results are underlined.}
		\vspace{-4mm}
		\label{table:sr}
	\end{table}
	Compared with the VoD dataset \cite{vod}, the Snail-Radar Dataset \cite{snailradar} contains longer trajectories and more complex environments, posing greater challenges to odometry stability. We compare our method with ICP-based point cloud odometry methods and 4DRadarSLAM \cite{4dradarslam}, a classical 4D radar SLAM method. We turn off the loop closure detection in 4DRadarSLAM to ensure a fair comparison. Additionally, we present the results of Full A-LOAM and A-LOAM without mapping, using 32-beam LiDAR point clouds as input. Table~\ref{table:sr} indicates that, while our method is slightly less effective than Full A-LOAM in long-range odometry localization, it narrows the performance gap with leading LiDAR odometry methods and significantly outperforms A-LOAM without mapping. Furthermore, compared with 4DRadarSLAM, our method reduces mean translation and rotation errors by 65.7\% and 71.6\%, respectively.
	
	\subsection{Ablation Study}
	\begin{table*}[t]
		%\centering
		\footnotesize
		\begin{center}
			\resizebox{1.0\textwidth}{!}
			{
				\begin{tabular}{l|l||cc|cc|cc|cc|cc|cc|cc||cc}
					\toprule
					& &  \multicolumn{2}{c|}{03}  &\multicolumn{2}{c|}{04}      & \multicolumn{2}{c|}{09} & \multicolumn{2}{c|}{17} &  \multicolumn{2}{c|}{19} & \multicolumn{2}{c|}{22} & \multicolumn{2}{c|}{24} &\multicolumn{2}{c}{Mean} \\ 
					\cline{3-18}\noalign{\smallskip}
					
					& \multirow{-2}{*}{\begin{tabular}[c]{@{}c@{}}Method \end{tabular}}
					&  $t_{rel}$  & $r_{rel}$   & $t_{rel}$                       & $r_{rel}$               & $t_{rel}$                          & $r_{rel}$   & $t_{rel}$ & $r_{rel}$   & $t_{rel}$                          & $r_{rel}$   & $t_{rel}$ & $r_{rel}$    & $t_{rel}$                          & $r_{rel}$   & $t_{rel}$ & $r_{rel}$      \\
					%\midrule
					\hline\hline
					%\cline{1-26}
					\noalign{\smallskip}
					(a)   
					&Ours w/o multi-scale geometric feature    
					&0.040	&0.067	&0.026	&0.033	&0.030	&\bf0.012	&0.017	&0.023	&\bf0.033	&0.150	&0.030	&0.049	&0.033	&0.033	&0.030	&0.052
					\\
					& Ours w/o clustering-based class-aware feature
					&0.043	&0.072	&0.025	&\bf0.030	&0.019	&\bf0.012	&0.017	&0.021	&0.050	&0.053	&0.023	&0.038	&0.031	&0.030	&0.030	&0.037
					\\
					& Ours w/o global transformer  
					&\bf0.032	&\bf0.037	&0.027	&0.031	&\bf0.016	&0.014	&0.018	&0.021	&0.045	&0.071	&0.027	&0.039	&\bf0.027	&\bf0.026	&0.027	&0.034
					\\
					
					&Ours (full)     
					&0.035 &0.056
					&\bf0.021 & 0.031
					&\bf0.016 & 0.018
					&\bf0.015 & \bf0.015
					&0.039 & \bf0.021
					&\bf0.016 & \bf0.028
					&\bf0.027 & 0.034
					&\bf0.024 & \bf0.029    
					\\
					\cline{1-18}\noalign{\smallskip}

					(b) &Ours w/o AMBA layer    
					&0.053	&0.115	&0.024	&0.085	&0.020	&0.053	&0.020	&0.048	&0.021	&0.159	&0.037	&0.076	&0.033	&0.089	&0.030	&0.089
					\\
					&Ours w/o iteration operator    
					&0.470	&0.814	&0.338	&0.701	&0.455	&0.704	&0.390	&0.442	&0.133	&0.021	&0.389	&0.600	&0.259	&0.873	&0.348	&0.594
					\\
					&Ours (full)     
					&\bf0.035 & \bf0.056
					&\bf0.021 & \bf0.031
					&\bf0.016 & \bf0.018
					&\bf0.015 & \bf0.015
					&\bf0.039 & \bf0.021
					&\bf0.016 & \bf0.028
					&\bf0.027 & \bf0.034
					&\bf0.024 & \bf0.029 
					\\
					\cline{1-18}\noalign{\smallskip}
					(c) 
					&Ours w/o confidence weight   
					&0.039	&\bf0.041	&0.027
					&0.040	&0.033	&0.027	&0.032	&0.025	&0.189	&0.259	&0.048	&0.043	&0.049	&0.109	&0.060	&0.078
					
					\\
					&Ours (one confidence weight)     
					&0.056	&0.092	&0.031	&0.039	&0.032	&0.049	&0.097	&\bf0.015	&0.065	&0.075	&0.024	&0.037	&0.037	&0.041	&0.049	&0.050
					
					\\
					&Ours (full)     
					&\bf0.035 & 0.056
					&\bf0.021 & \bf0.031
					&\bf0.016 & \bf0.018
					&\bf0.015 & \bf0.015
					&\bf0.039 & \bf0.021
					&\bf0.016 & \bf0.028
					&\bf0.027 & \bf0.034
					&\bf0.024 & \bf0.029    
					\\
					\bottomrule
				\end{tabular}
			}
		\end{center}
		\caption{The ablation study results of 4D Radar odometry on the VoD dataset \cite{vod}. The best results are bold.}		
		\label{table:ablation}
	\end{table*}
	\begin{figure*}[t]
		\centering
		\begin{subfigure}{0.46\linewidth}
			\includegraphics[width=\linewidth]{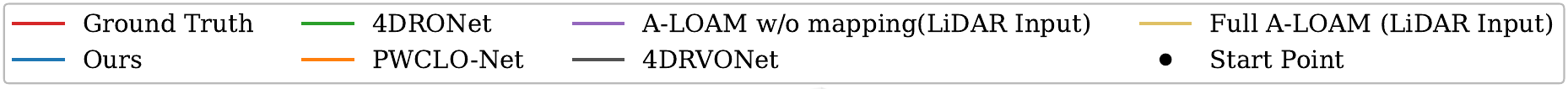}
		\end{subfigure}
		\hfill
		\begin{subfigure}{0.43\linewidth}
			\includegraphics[width=\linewidth]{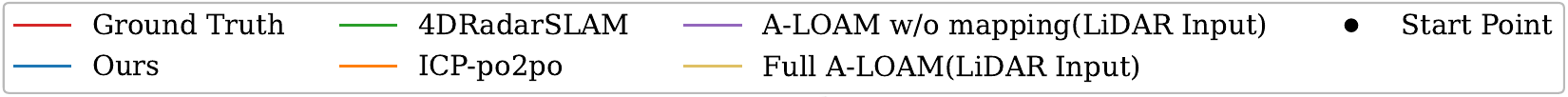}
		\end{subfigure}
		
		\begin{subfigure}{0.22\linewidth}
			\includegraphics[width=\linewidth]{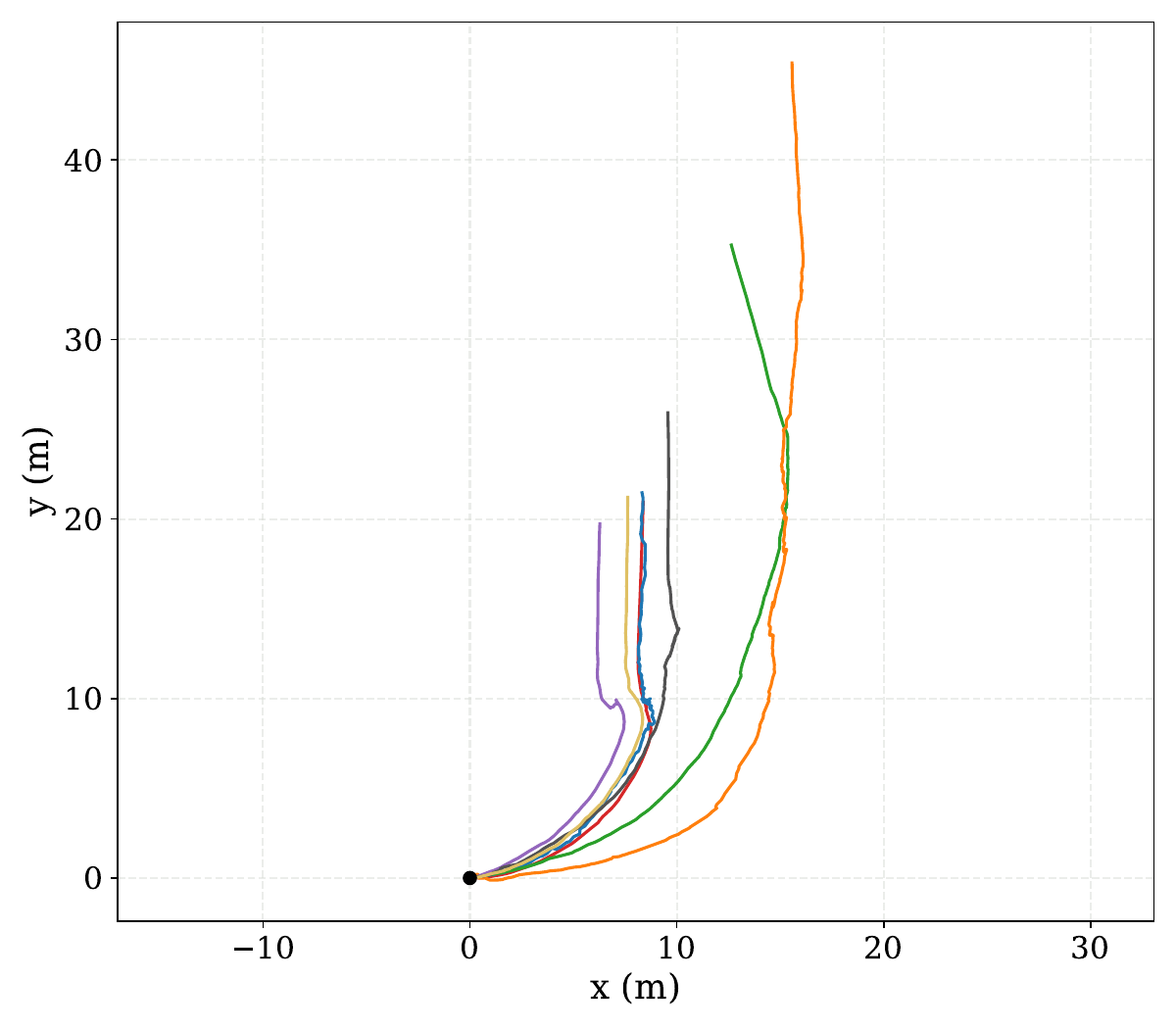}
		\end{subfigure}
		\hfill
		\begin{subfigure}{0.22\linewidth}
			\includegraphics[width=\linewidth]{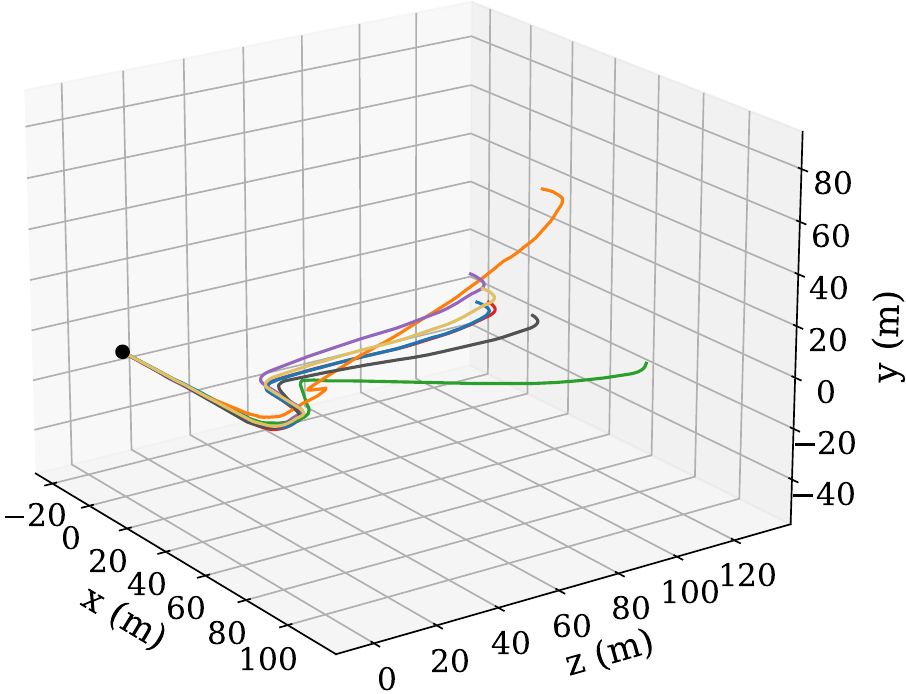}
		\end{subfigure}
		\hfill
		\begin{subfigure}{0.22\linewidth}
			\includegraphics[width=\linewidth]{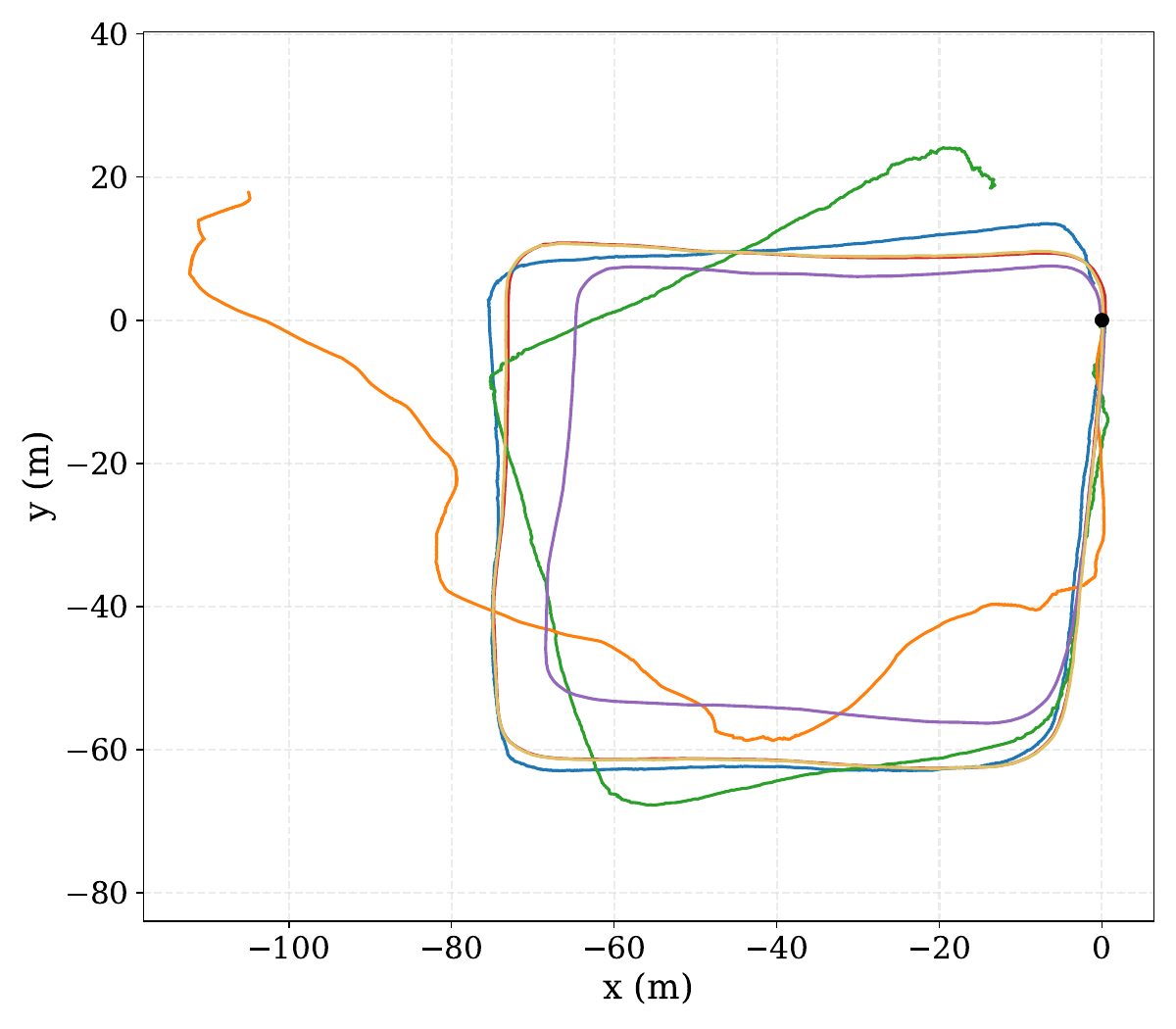}
		\end{subfigure}
		\hfill
		\begin{subfigure}{0.22\linewidth}
			\includegraphics[width=\linewidth]{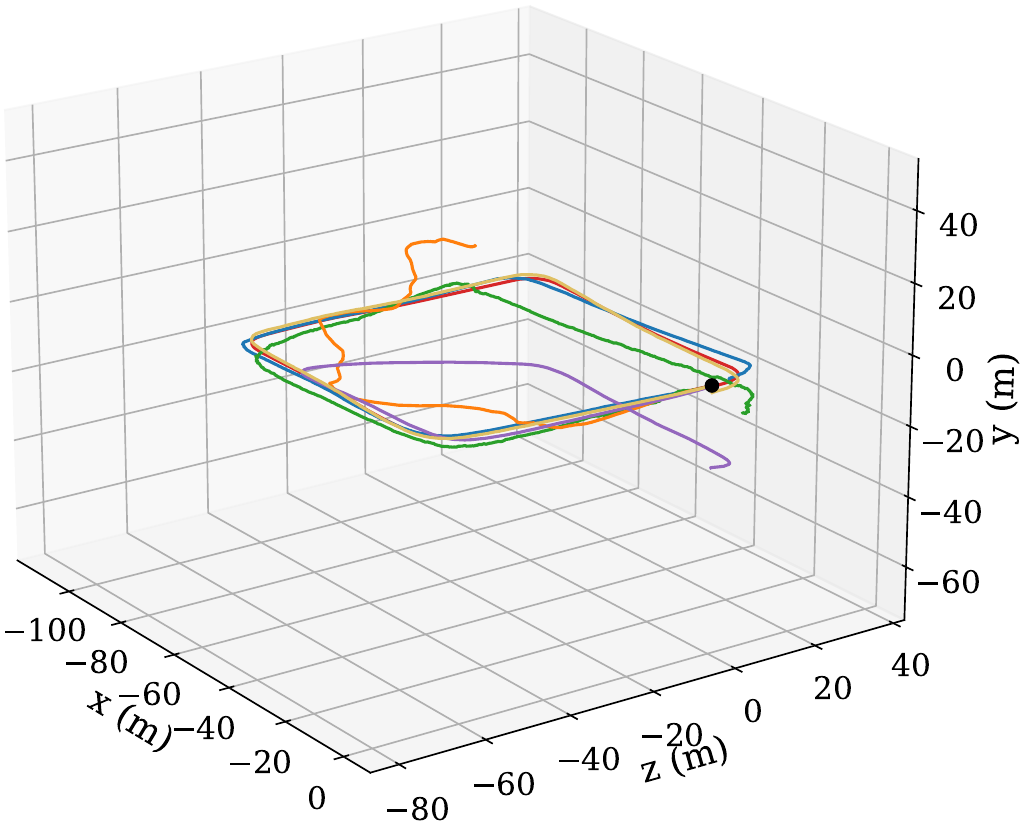}
		\end{subfigure}
		\caption{3D and 2D trajectory results for VoD test sequences 19 and 22, and Snail-Radar test sequences st. Our method obtains the most accurate trajectory.}
		\label{fig4}
	\end{figure*}
	
	\begin{figure}[t]
		\centering
		\begin{subfigure}{0.98\linewidth}
			\centering
			\includegraphics[width=\linewidth]{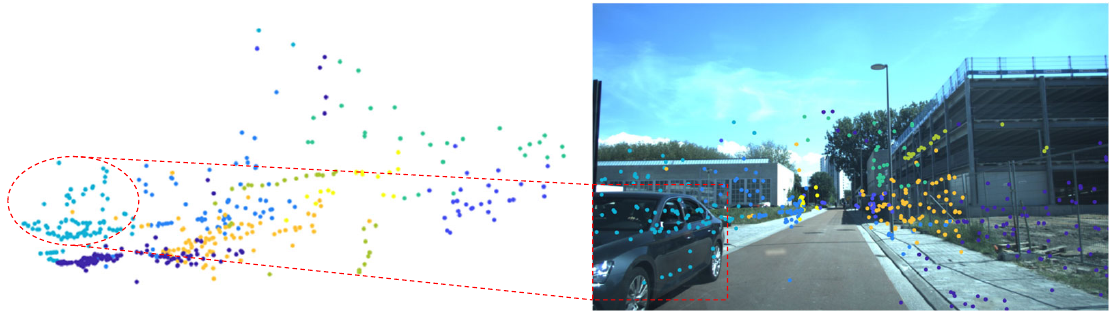}
			\label{fig5a}
		\end{subfigure}
		\caption{Visualization of clustering results. 4D radar points are color-coded based on their assigned categories.}
		\label{fig5}
	\end{figure}
	We perform ablation experiments on the VoD Dataset to evaluate the effectiveness of each module in DNOI-4DRO.
	
	\noindent{}{\bf Feature Extraction:} We evaluate the effectiveness of each component in the proposed feature extraction network. As shown in Table~\ref{table:ablation}(a), removing any individual module leads to a noticeable drop in pose estimation accuracy, highlighting each module's critical role in capturing fine-grained 4D radar features and enhancing the overall 4DRO performance.
	
	\noindent{}{\bf Iteration Operator:} We remove the AMBA layer during training, relying on point motion flow supervision to train the network. Then, we eliminate the entire iteration operator and directly regress poses from correlation features, which means that completely discarding the geometric optimization process. As shown in Table~\ref{table:ablation}(b), removing either component significantly reduces pose estimation accuracy. These results highlight the crucial role of the introduced geometric optimization mechanism in enhancing estimation precision.
	
	\noindent{}{\bf Confidence Weight:} First, we remove the confidence weight by setting all weights to 1. Then, we calculate a confidence value for each point, assuming that the point has equal importance in pose estimation across different directions. Table~\ref{table:ablation}(c) show that the proposed confidence weight and fine-grained estimation across different directions contribute to better results.
	
	\subsection{Visualization}
	We visualize 3D and 2D trajectories based on our estimated pose on the VoD and Snail-Radar datasets, as shown in Figure~\ref{fig4}. These figures show that our odometry can track the trajectory of the ground truth fairly well. Although we do not have the mapping procedure, our odometry achieves trajectory accuracy comparable to that of Full A-LOAM in short-range localization. Furthermore, our approach significantly reduces odometry drift along the Z-axis compared to other methods. Figure~\ref{fig5} illustrates the clustering results. The clustering-based method effectively mitigates the constraints of the receptive field, facilitating extracting features that belong to the same category (e.g., vehicles, curbs, walls). This approach enables point features to exhibit class awareness, improving their representational capacity.
	
	\section{Conclusion}
	\label{sec:conclusion}
	We introduce DNOI-4DRO, an end-to-end neural architecture for 4D radar odometry, which combines the strengths of both classical approaches and deep networks through a differentiable neural-optimization iteration module. We compare against existing approaches and show strong performance across several datasets.
	
	\section{Acknowledgements}
	\label{sec:acknowledgements}
	This work was supported by the National Natural Science Foundation of China under Grant 52325212.
	
	\bibliography{aaai2026}

\end{document}